\documentclass[11pt]{article}

\usepackage[preprint]{acl}
\usepackage{stfloats}   
\usepackage[table]{xcolor}
\usepackage{subcaption}
\usepackage{colortbl}
\usepackage{booktabs}
\usepackage{times}
\usepackage{multirow}
\usepackage{tabularx}
\usepackage{latexsym}
\usepackage{placeins} 
\usepackage[T1]{fontenc}

\usepackage[utf8]{inputenc}

\usepackage{microtype}

\usepackage{inconsolata}
\usepackage{cuted}
\usepackage{graphicx}

\usepackage{amsmath}    
\usepackage{amssymb}    
\usepackage{amsfonts}   
\usepackage{mathtools}  
\usepackage{bm}

\title{Sink-Aware Pruning for Diffusion Language Models}

\author{Aidar Myrzakhan, ~Tianyi Li, ~Bowei Guo, ~Shengkun Tang, ~Zhiqiang Shen \\
  VILA Lab, MBZUAI \\
  }

\begin{document}
\maketitle
\begin{abstract}
Diffusion Language Models (DLMs) incur high inference cost due to iterative denoising, motivating efficient pruning. Existing pruning heuristics largely inherited from autoregressive (AR) LLMs, typically preserve attention sink tokens because AR sinks serve as stable global anchors. We show that this assumption does not hold for DLMs: the attention-sink position exhibits substantially higher variance over the full generation trajectory (measured by how the dominant sink locations shift across timesteps), indicating that sinks are often transient and less structurally essential than in AR models. Based on this observation, we propose {\bf \texttt{Sink-Aware Pruning}}, which automatically identifies and prunes unstable sinks in DLMs (prior studies usually keep sinks for AR LLMs).
Without retraining, our method achieves a better quality-efficiency trade-off and outperforms strong prior pruning baselines under matched compute. Our code is available at \url{https://github.com/VILA-Lab/Sink-Aware-Pruning}.
\end{abstract}

\section{Introduction}

Diffusion Language Models (DLMs)~\cite{li2025survey, nie2025large, ye2025dream, zhu2025llada} generate text via iterative denoising over multiple timesteps, repeatedly updating the full token sequence (or a latent representation) until it converges. This contrasts with autoregressive (AR) LLMs~\cite{radford2018improving,radford2019language}, which generate tokens one-by-one with a single forward pass per new token. While DLMs recently can achieve attractive generation properties, their iterative inference substantially increases compute and memory cost, making many acceleration, especially pruning, critical for practical deployment. Most existing pruning recipes, however, are adapted from AR Transformers and implicitly assume that attention behaviors (and their {\em important tokens}) transfer unchanged to diffusion-style generation.

\begin{figure}[t]
  \centering
  \hspace*{-0.02\textwidth}
  \includegraphics[width=0.5\textwidth]{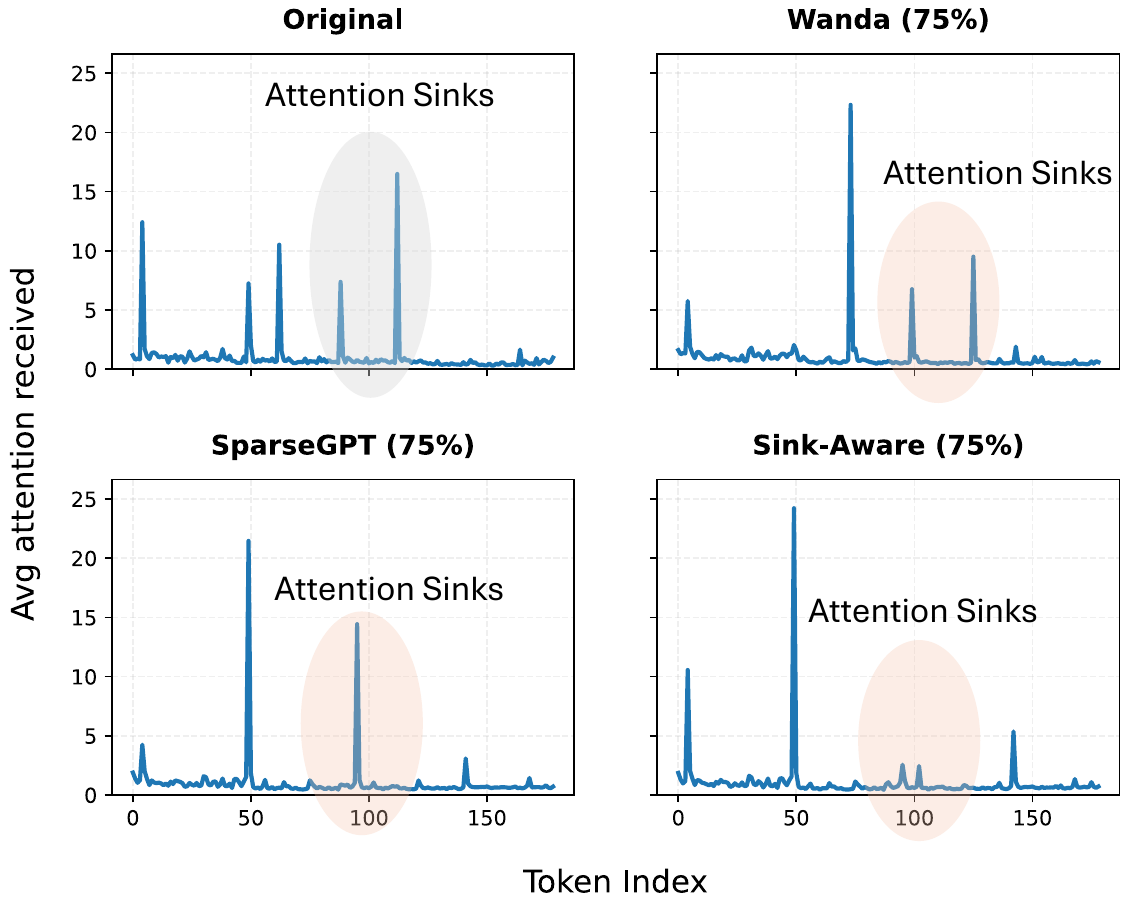}
  \vspace{-0.3in}
  \caption{Illustration of attention sink behaviors in Diffusion Language Models. Ours {\texttt{Sink-Aware Pruning}} reduces sink variance by downscaling unstable sinks.}
 \label{sink_score}
\end{figure}

A key AR-specific phenomenon is the existence of attention sink tokens: a small set of positions (often early tokens such as {\em BOS}, {\em system prompts}, or a few {\em prefix tokens}) that consistently attract disproportionately large attention mass across layers and heads. In AR models, these sinks behave as stable global anchors that help propagate conditioning information and stabilize residual-stream dynamics across the long, causal computation graph in autoregressive-based attention. Consequently, many AR pruning and token-dropping methods explicitly preserve sinks (or protect early/prefix tokens) to avoid catastrophic quality degradation. This practice has become a de facto heuristic: when pruning attention structure, do not remove sink positions because they are assumed to be universally important in AR models.

\begin{figure*}[h]
  \centering
  \hspace*{0.02\textwidth}
  \includegraphics[width=\textwidth]{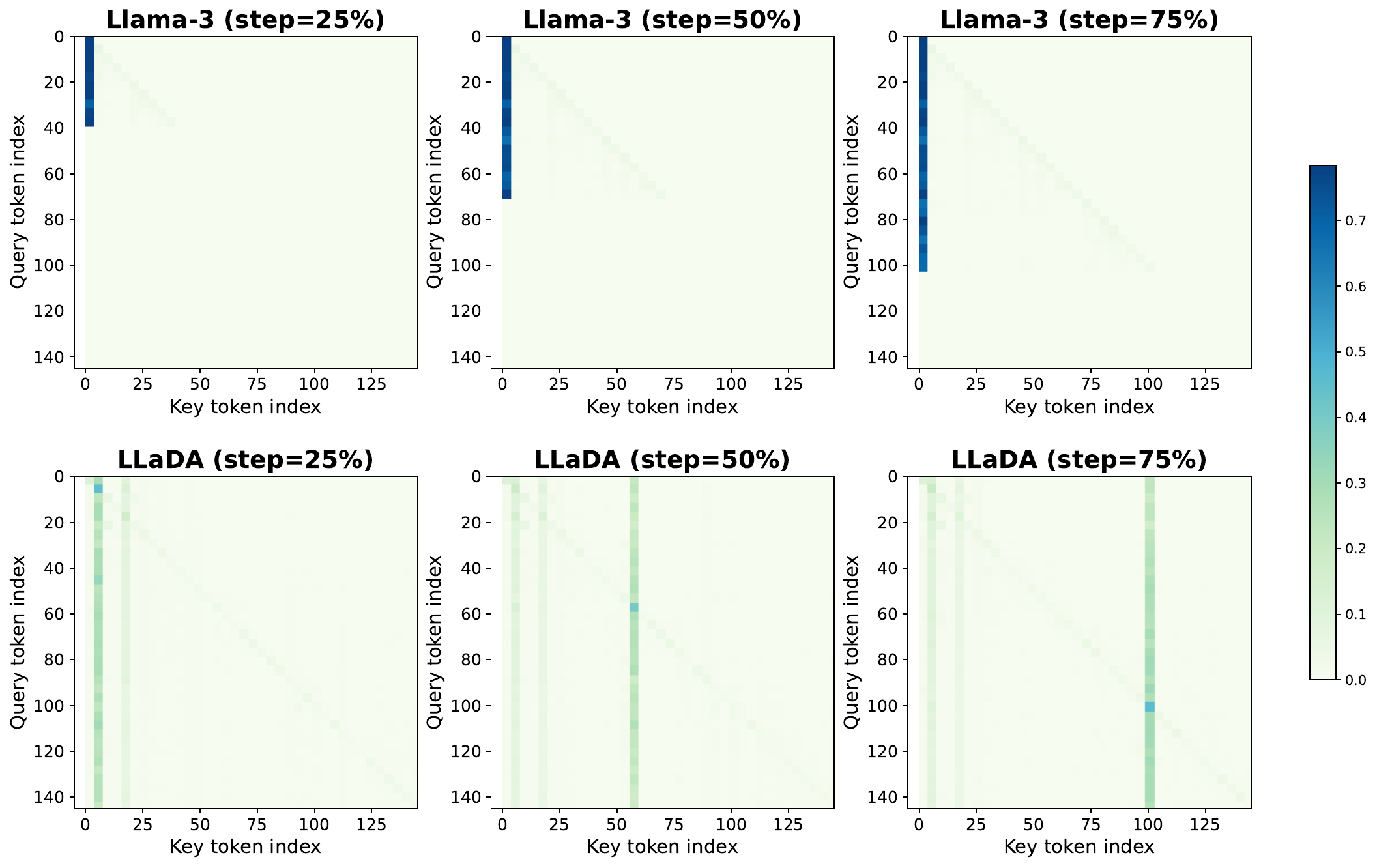}
  \vspace{-0.3in}
  \caption{Attention sink heatmap dynamics across generation steps for AR LLM (LLaMA-3-8B) and DLM (LLaDA). For each model,
  we show 3 different generation stages (25, 50, and 75\% of the total generation steps) and plot the attention mass received by each token position (y-axis) across all heads/layers (x-axis). 
  In LLaMA, the sink position (deep-blue vertical band) is stable across steps, while in LLaDA, the sink position shifts significantly across diffusion steps, indicating higher sink variance. The step in AR model refers to the generation process.}
  \label{fig:attention_sink}
\end{figure*}

We revisit this assumption for DLMs and find it does not hold. Because DLMs update all tokens at each diffusion timestep, the model's attention organization evolves throughout the denoising trajectory: early steps must resolve global structure under high noise, while later steps refine local syntax and semantics under lower noise. To quantify sink stability, we compute attention statistics at every generation step for every token: for AR models, ``step'' corresponds to each newly generated token; for DLMs, ``step'' corresponds to each diffusion timestep (with attention computed over the whole sequence each time). We then track the sink position (e.g., the token index receiving maximal aggregated attention, or the top-$k$ sink indices) across the entire generation process and define sink variance as the degree to which these sink positions shift over steps, as shown in Fig.~\ref{fig:attention_sink}.

This analysis (as shown in Fig.~\ref{sink_score} and Fig.~\ref{fig:temporal_sink}) reveals a clear divergence: sink-token variance in DLMs is substantially larger than in AR LLMs. In AR generation, sink positions tend to be highly persistent: once a sink emerges (typically near the prefix), it remains a sink across subsequent token emissions and across many heads / layers. In DLMs, by contrast, the identity and location of the dominant sink frequently changes across diffusion timesteps, reflecting the model's shifting needs as denoising progresses. Many sinks in DLMs are therefore transient: they may attract attention at some timesteps (often under high noise or during global structure formation) but lose prominence later, and different sinks may take over as the sequence becomes more refined. This implies that {\em always keep sinks} is not a diffusion-invariant principle, rather, it is a property of causal, prefix-conditioned AR computation.

Motivated by this, we propose {\bf \texttt{Sink-Aware Pruning}}: a diffusion-specific pruning strategy that distinguishes stable anchors from ephemeral sinks, and prunes accordingly. Instead of hard-coding sink preservation, we (i) estimate sink variance over the denoising trajectory, (ii) identify unstable sink candidates whose sink positions fluctuate significantly across diffusion timesteps, and (iii) prune these unstable sinks to reduce attention / weight cost while preserving informative pathways. 
Crucially, the method is generation-paradigm aware: for AR LLMs, where sink variance is low and sinks are structurally important, thus prior work keeps sinks, for DLMs in this work, where sink variance is high, we allow and encourage sink pruning. 

Empirically, {\bf \texttt{Sink-Aware Pruning}} improves the quality-efficiency trade-off over strong prior pruning baselines under matched compute. Across multiple DLM settings, pruning transient sinks reduces redundant or global attention that does not persist across timesteps, enabling more aggressive pruning without the failure modes observed when AR-centric sink-preservation heuristics are naively applied to DLMs. The results support our central claim: attention sinks are not universally {\em must-keep} tokens, their utility depends on the generation dynamics. By explicitly measuring sink-position variance and tailoring pruning decisions to diffusion timesteps, Sink-Aware Pruning provides a principled and effective route to accelerating DLM inference without retraining.

In summary, our contributions are threefold:
\begin{itemize}
\item We introduce sink-position variance to track how attention-sink indices shift over the full generation trajectory, and show it is much higher in diffusion LLMs than in autoregressive LLMs when computing attention at every AR step / diffusion timestep for each token.
\item We demonstrate that the AR heuristic {\em always keep attention sinks} does not transfer to diffusion generation: sinks are stable anchors in AR models but are often transient in DLMs, so they can be pruned.
\item We propose {\bf \texttt{Sink-Aware Pruning}}, which prunes unstable sinks in DLMs (while conventional AR LLMs preserve sinks) with timestep-aware retention of salient attentions, and outperforms strong prior pruning baselines under the same compute.
\end{itemize}

\section{Related Work}
\subsection{Diffusion Language Models}
Diffusion Language Models (DLMs)~\citep{li2025survey,nie2025large,ye2025dream} have  recently emerged as a promising non-autoregressive alternative to conventional autoregressive language models. Inspired by diffusion models in continuous domains~\cite{ho2020denoising,rombach2022high}, DLMs formulate text generation as an iterative denoising process, enabling parallel generation and naturally incorporating bidirectional context. Existing work broadly categorizes DLMs into continuous-space and discrete-space formulations. Continuous DLMs operate on token embeddings or logits and apply Gaussian or flow-based diffusion processes~\citep{gong2022diffuseq,li2022diffusion}, while discrete DLMs define diffusion directly over categorical token spaces using masking or structured transition operators~\citep{austin2021structured,he2023diffusionbert,sahoo2024simple}. Recent advances demonstrate that large-scale discrete DLMs can achieve 
performance comparable to strong autoregressive baselines while offering substantial inference speedups through parallel decoding~\citep{wu2025fast, ni2025diffusion}. Moreover, hybrid autoregressive-diffusion approaches further balance generation quality and efficiency by combining block-wise autoregression with intra-block diffusion refinement~\citep{arriola2025block, cheng2025sdar}.

\subsection{Attention Sink in LLMs}

Initially identified in LLMs by~\citet{xiao2023efficient}, attention sinks are a small set of tokens (often early in the context) that absorb disproportionate attention across many layers / heads despite limited semantic value~\citep{gu2024attention, barbero2025llms}. \citet{xiao2023efficient} attribute sinks to Softmax normalization: when a query has no strong match, attention mass must still be assigned somewhere, and globally visible early tokens become a natural {\em dumping ground} for redundant attention.

Recent work shows sinks also appear in masked DLMs but with different behavior~\citep{rulli2025attention}. Studying large masked DLMs (e.g., LLaDA~\citep{nie2025large}, Dream~\citep{ye2025dream}, MMaDA~\citep{yang2025mmada}), they find sink locations are step-dependent, emerging, shifting, or vanishing as denoising progresses. Moreover, unlike autoregressive models where removing sinks can severely hurt performance, masked DLMs are relatively robust: masking top sinks during generation causes only minor drops. This suggests bidirectional iterative denoising provides alternative aggregation pathways, making sink dynamics central to long-context efficiency and stability across decoding paradigms.

\begin{figure*}[t]
  \centering
  \includegraphics[width=0.95\textwidth]{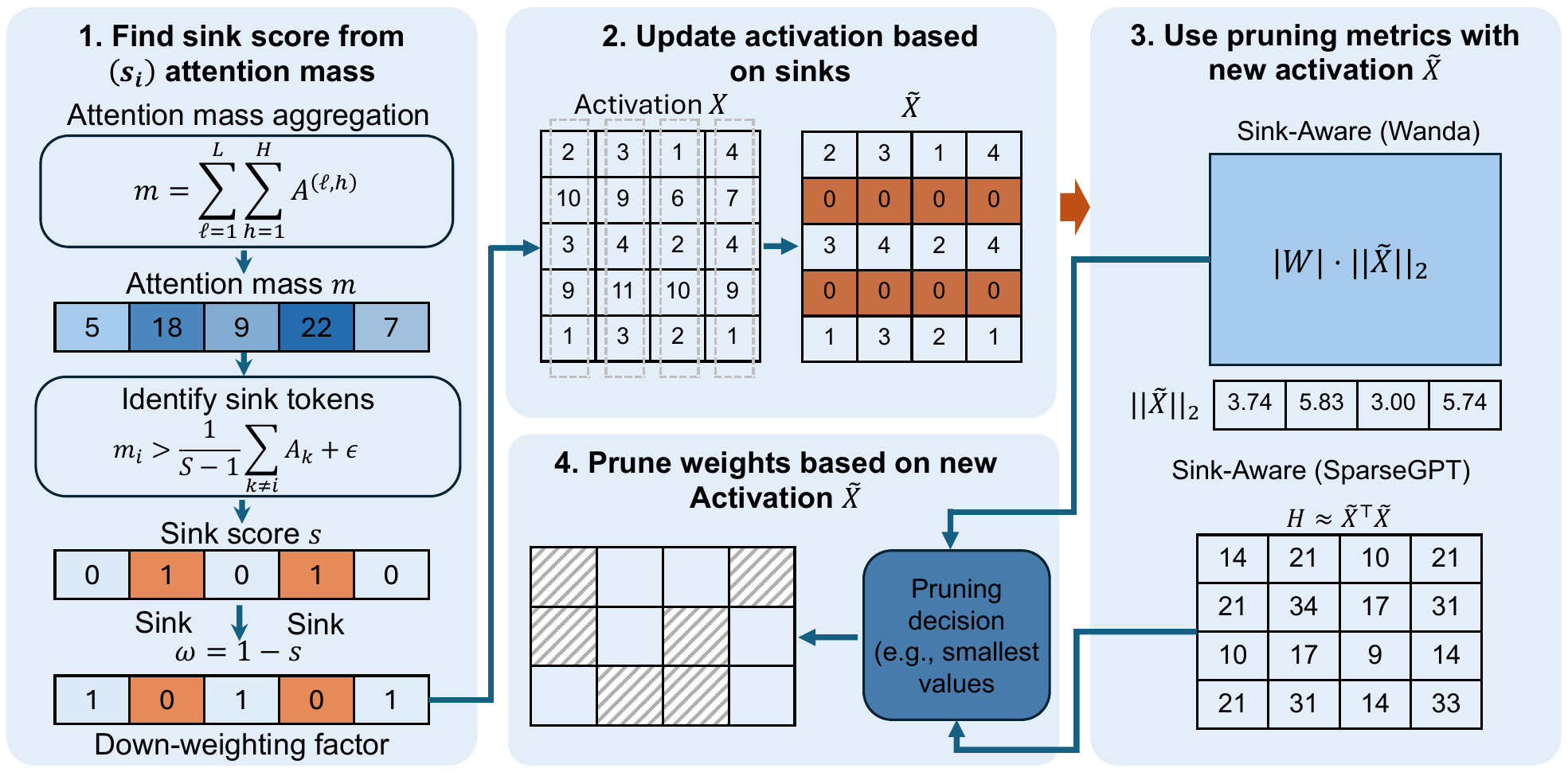}
  \caption{\textbf{Overview of Sink-Aware Pruning}. Given input activations, we compute per-token attention mass aggregated across all layers and heads (Step 1), identify sink tokens via a threshold-based criterion, and derive a soft down-weighting factor $\omega = 1-s$. The original activation $X$ is then suppressed at sink positions to produce a new activation $\tilde{X}$ (Step 2), which is substituted into existing pruning criteria, Wanda or SparseGPT, to compute sink-aware importance scores (Step 3). Final pruning decisions are made based on the updated scores (Step 4).}
  \label{fig:sink}
\end{figure*}

\subsection{LLM Pruning}
Network pruning has long been studied as a model compression technique that reduces inference cost by removing redundant parameters based on importance criteria~\citep{lecun1989optimal,hassibi1993optimal,han2015deep,cheng2024survey}. Existing approaches are commonly categorized into unstructured and 
structured pruning. While unstructured pruning offers fine-grained flexibility, structured pruning removes entire neurons, channels, or matrix rows/columns, making it more amenable to hardware acceleration but often at the cost of higher accuracy degradation~\citep{han2015learning,liu2017learning,molchanov2019importance}.

As large language models continue to scale, applying traditional pruning pipelines becomes increasingly challenging due to the prohibitive cost of retraining. This has motivated a shift toward post-training pruning methods, which aim to identify and remove unimportant weights directly from pretrained models using lightweight importance metrics, without full retraining~\citep{sun2023simple,frantar2023sparsegpt,das2023beyond,yang2025wanda++}. These approaches have demonstrated promising results for compressing LLMs with minimal performance loss, and highlight pruning as a practical tool for improving inference efficiency in large-scale models~\citep{ma2023llm}.
\section{{\bf \texttt{Sink-Aware Pruning}} for DLMs}
\subsection{Preliminaries}
\paragraph{Wanda.}
Wanda~\citep{sun2023simple} assigns each weight an output-wise importance score that combines its magnitude with the norm of the corresponding input activation.
For a linear layer with weights $W \in \mathbb{R}^{C_{\text{out}} \times C_{\text{in}}}$ and input activations $X \in \mathbb{R}^{N \times C_{\text{in}}}$ collected from a calibration set, the importance score of weight $W_{ij}$ is defined as:
\begin{equation}
S_{ij} = |W_{ij}| \cdot \lVert X_{\cdot j} \rVert_2.
\label{eq:wanda_score}
\end{equation}
where $X_{\cdot j} \in \mathbb{R}^{N}$ denotes the $j$-th input feature across samples.
Pruning is performed independently for each output neuron by comparing scores within each row group $G_i = \{W_{i,1}, \ldots, W_{i,C_{\text{in}}}\}$ and removing the lowest $s\%$ weights per output.
The activation norms $\lVert X_{\cdot j} \rVert_2$ are estimated without gradient or Hessian computation, making Wanda computationally efficient for large models.

\paragraph{SparseGPT.}
SparseGPT~\citep{frantar2023sparsegpt} formulates the post-training pruning process as a layer-wise reconstruction problem using second-order information.
For a layer with original weights $W$ and layer inputs $X$, the pruning objective is defined as:
\begin{equation}
\min_{M,\widetilde W}\; \lVert W X - (M \odot \widetilde W) X \rVert_F^2,
\label{eq:sparsegpt_obj}
\end{equation}
where $M$ is a binary pruning mask and $\widetilde W$ denotes the reconstructed weights.
Given a fixed mask $M_i$ for the $i$-th output row, the optimal remaining weights admit a closed-form linear regression solution:
\begin{equation}
\mathbf{w}^i_{M_i}
=
\left(
\mathbf{X}_{M_i}\mathbf{X}_{M_i}^{\top}
\right)^{-1}
\mathbf{X}_{M_i}
\left(
\mathbf{w}_{M_i}\mathbf{X}_{M_i}
\right)^{\top},
\label{eq:sparsegpt_lstsq}
\end{equation}
Based on a second-order approximation, the loss increase induced by removing a scalar weight $w_m$ is estimated as:
\begin{equation}
\varepsilon_m \;=\; \frac{w_m^2}{\left[H^{-1}\right]_{mm}}.
\label{eq:sparsegpt_score}
\end{equation}
where $H = X X^\top + \lambda I$ is the empirical Hessian approximation.
SparseGPT iteratively prunes weights with small $\varepsilon_m$ and updates the remaining weights to reduce reconstruction error, enabling high sparsity without retraining.

\subsection{Sink-Aware Pruning}

\paragraph{Variance Statistics in DLMs and AR models.}
To characterize how attention sinks differ between diffusion and autoregressive language models, we introduce two complementary variance statistics. Let $\mathbf{A}^{(t)} \in \mathbb{R}^{N \times N}$ denote the attention matrix at generation step $t$ (a diffusion timestep for DLMs; a token emission for AR models), where $N$ is the sequence length. We define the \textit{incoming attention mass} at position $i$ and step $t$ as:
\begin{equation}
    m_t(i) = \sum_{j=1}^{N} A^{(t)}_{j,i},
\end{equation}
which measures how much total attention all other tokens direct toward position $i$. Positions receiving substantially more incoming attention than the average act as attention sinks at that step.

We capture two dimensions of sink behavior. \textbf{Spatial variance} measures how unevenly attention concentrates across positions when averaged over the full trajectory. We compute $\bar{m}(i) = \frac{1}{T}\sum_{t=1}^{T} m_t(i)$ and take
\begin{equation}
    \sigma^2_{\text{spatial}} = \mathrm{Var}_{i}\!\left(\bar{m}(i)\right).
\end{equation}
A large value indicates that a few positions dominate the attention landscape on average, but does not reveal whether those positions stay fixed over time. \textbf{Temporal variance} captures how much the sink location shifts across steps. At each step $t$, we compute the attention-weighted centroid over the current sink set $\mathcal{S}_t$ (defined below):
\begin{equation}
    c_t = \frac{\sum_{i \in \mathcal{S}_t} m_t(i) \cdot i}{\sum_{i \in \mathcal{S}_t} m_t(i)}, \qquad \sigma^2_{\text{temporal}} = \mathrm{Var}_{t}(c_t).
\end{equation}
A near-zero temporal variance means the sink remains locked to the same region throughout generation, while a large value indicates substantial migration.

\begin{figure}[t]
  \centering
  \begin{subfigure}{0.45\textwidth}
    \centering
    \includegraphics[width=\linewidth]{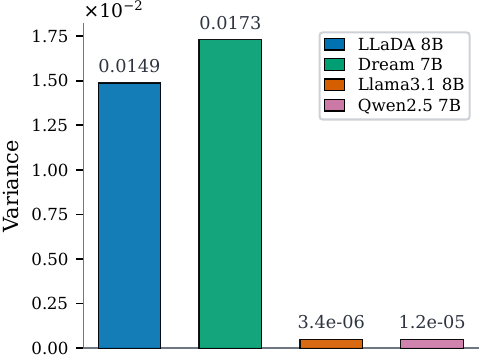}
    \caption{Variance of the sink across generation/denoising steps.}
    \label{fig:temporal_sink}
  \end{subfigure}
  \hfill
  \begin{subfigure}{0.45\textwidth}
    \centering
    \includegraphics[width=\linewidth]{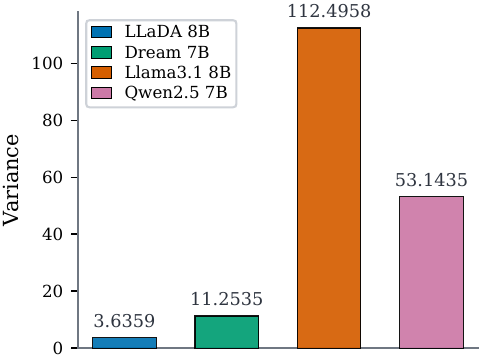}
    \caption{Variance of total attention received by each token across all the generation/denoising steps.}
    \label{fig:spatial_sink}
  \end{subfigure}
  \vspace{-0.1in}
  \caption{Attention sink variance for diffusion LMs (LLaDA, Dream) and autoregressive LMs (Llama~3.1, Qwen~2.5).}
  \label{fig:sink_variance}
\end{figure}

\begin{figure}[t]
  \centering
  \includegraphics[width=0.5\textwidth]{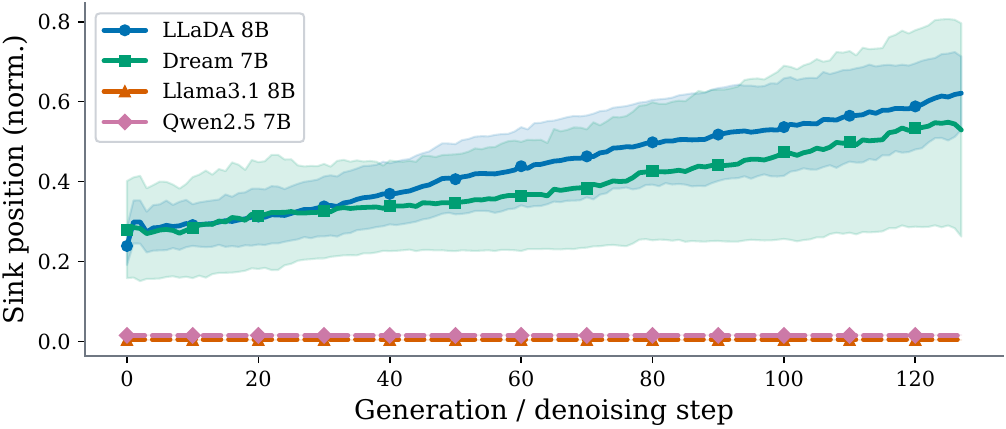}
  \caption{Sink position across generation/denoising steps for diffusion and AR LMs. Shaded regions denote $\pm$ std across runs.}
  \label{fig:sink_trajectory}
\end{figure}

As shown in Fig.~\ref{fig:sink_variance}, these two statistics reveal a clear contrast. AR models exhibit high spatial variance but near-zero temporal variance: their sinks are concentrated on a small, fixed set of early positions and remain stationary across all generation steps. DLMs show the opposite pattern: lower spatial variance (attention is more distributed) but temporal variance that is orders of magnitude larger. The sink trajectory (Fig.~\ref{fig:sink_trajectory}) provides further intuition: for AR models, the sink stays flat, while for DLMs it drifts progressively from earlier to later positions as denoising advances, with wide step-to-step variability. This drift reflects how early diffusion steps focus on global structure under high noise, while later steps shift attention toward local, token-level refinement. In short, AR sinks are spatially concentrated and temporally stable, while DLM sinks are spatially diffuse and temporally transient.

\vspace{-0.20cm}
\paragraph{Pruning Metric.}
We now formalize the notion of attention sinks used in our pruning criterion. We begin by computing the \emph{attention mass} received by each token position. At denoising step~$t$, the cumulative attention score of token~$j$ in head~$h$ of layer~$\ell$ is:
\begin{equation}
  \bar{A}_j^{(t,\ell,h)} \;=\; \frac{1}{S}\sum_{i=1}^{S} A_{i,j}^{(t,\ell,h)},
  \label{eq:cum_attn}
\end{equation}
where $A_{i,j}^{(t,\ell,h)}$ is the attention weight from query position~$i$ to key position~$j$ and $S$ is the sequence length.
We aggregate across all layers and heads to obtain a position-level attention mass:
\begin{equation}
  m_t(j) \;=\; \sum_{\ell=1}^{L}\sum_{h=1}^{H} \bar{A}_j^{(t,\ell,h)},
  \label{eq:attn_mass}
\end{equation}
A token~$j$ is considered a \textbf{sink token} at step~$t$ if its attention mass significantly exceeds that of other positions, i.e.,
\begin{equation}
  m_t(j) \;>\; \frac{1}{S-1}\sum_{k \neq j} m_t(k) \;+\; \epsilon,
  \label{eq:sink_def}
\end{equation}
where $\epsilon > 0$ controls the detection sensitivity.
To obtain a differentiable relaxation, we convert this criterion into a soft sink score via a sigmoid:
\begin{equation}
  \phi_t(j) \;=\; \sigma\!\!\left(\!(m_t(j) - \frac{1}{S-1}\sum_{k \neq j} m_t(k) - \epsilon\right),
  \label{eq:soft_sink}
\end{equation}
We evaluate~\eqref{eq:soft_sink} on noised calibration inputs at a uniformly spaced set of timesteps $\mathcal{T} \subset \{1,\dots,T\}$ and average to obtain a step-invariant sink estimate:
\begin{equation}
  \bar{\phi}(k) \;=\; \frac{1}{|\mathcal{T}|}\sum_{t \in \mathcal{T}} \phi_t(k).
  \label{eq:avg_sink}
\end{equation}

\paragraph{Sink-Aware Importance Reweighting.}
Given the soft sink score $\bar{\phi}(j)$, we define a per-position weight
\begin{equation}
    \omega_j = 1 - \bar{\phi}(j),
\end{equation}
which is near zero for sink tokens and near one elsewhere.
We construct a \emph{sink-masked activation} $\widetilde{\mathbf{X}}$ by scaling each row of the original activation $\mathbf{X} \in \mathbb{R}^{S \times C_{\mathrm{in}}}$ by $\omega_j$, effectively suppressing the contribution of sink positions:
\begin{equation}
  \widetilde{\mathbf{X}}_{j,:} \;=\; \omega_j \cdot \mathbf{X}_{j,:},
  \qquad j = 1,\dots,S.
  \label{eq:sink_mask_input}
\end{equation}
This masked activation replaces $\mathbf{X}$ in existing pruning criteria.
For Wanda base method, we substitute $\widetilde{\mathbf{X}}$ into Eq.~\eqref{eq:wanda_score} yields the sink-aware score $\widetilde{S}_{ij} = |W_{ij}| \cdot \lVert \widetilde{X}_{\cdot\, j} \rVert_2$, and pruning proceeds as before.
For SparseGPT as the base approach, the empirical Hessian is computed from sink-masked inputs, $\widetilde{H} = \frac{1}{|\mathcal{D}|}\sum_{n} \widetilde{\mathbf{X}}_n^\top \widetilde{\mathbf{X}}_n + \lambda \mathbf{I}$, where $\mathcal{D}$ is the calibration dataset, with the pruning and reconstruction procedure following Eqs.~\eqref{eq:sparsegpt_obj}--\eqref{eq:sparsegpt_score} identically.
By suppressing sink tokens in the Hessian, the reconstruction allocates capacity toward faithfully reproducing the layer's output on semantically meaningful positions, rather than preserving behavior on tokens whose outsized activations would otherwise dominate the second-order statistics.

\section{Experiments}

\subsection{Experimental Setup}
\paragraph{Models and Benchmarks.}
We evaluate our sink-aware pruning method on multiple widely adopted pretrained diffusion language models (DLMs), including LLaDA~\citep{nie2025large}, Dream~\citep{ye2025dream}, and LLaDA-1.5~\citep{zhu2025llada}, as well as the multimodal unified DLM MMaDA~\citep{yang2025mmada}. For text generation and reasoning tasks, we assess the pruned models on a diverse set of standard language model benchmarks: MMLU~\citep{hendrycks2020measuring} (5-shot), ARC-C~\citep{clark2018think} (0-shot), PIQA~\citep{bisk2020piqa} (0-shot), WinoGrande~\citep{sakaguchi2021winogrande} (0-shot), HellaSwag~\citep{zellers2019hellaswag} (0-shot), RACE~\citep{lai-etal-2017-race} (0-shot), GSM8K~\citep{cobbe2021training} (5-shot), and GPQA~\citep{rein2024gpqa} (5-shot).

\begin{table*}[t]
\centering
\small
\setlength{\tabcolsep}{3.5pt}
\renewcommand{\arraystretch}{1.08}
\resizebox{0.87\textwidth}{!}{
\begin{tabular}{c l >{\columncolor{gray!20}}c c c c c c c c c}
\toprule
\textbf{Sparsity} & \textbf{Method} & \cellcolor{gray!20}\textbf{Avg.} & \textbf{MMLU} & \textbf{ARC-C} & \textbf{PIQA} & \textbf{WinoG} & \textbf{GSM8K} & \textbf{HellaSwag} & \textbf{GPQA} & \textbf{RACE} \\
\midrule
\multirow{1}{*}{Dense} & \textbf{Base} & 57.93 & 65.97 & 43.00 & 74.10 & 69.30 & 69.29 & 72.70 & 30.40 & 38.70 \\
\midrule
\multirow{4}{*}[-1.1ex]{0.25} & Wanda & 57.43 & 65.20 & \textbf{43.94} & \textbf{75.30} & \textbf{68.59} & 66.03 & 71.95 & \textbf{29.85} & \textbf{38.55} \\
\rowcolor{blue!5} & {\bf \texttt{Sink-Aware}}  & \cellcolor{gray!20}\textbf{57.62} & \textbf{65.41} & 43.52 & 74.97 & \textbf{68.59} & \textbf{68.16} & \textbf{72.30} & 29.70 & 38.32 \\
\cmidrule(lr){2-11}
& SparseGPT & 57.23 & 65.16 & \textbf{43.09} & \textbf{74.43} & 67.56 & 67.17 & \textbf{72.10} & \textbf{30.25} & 38.10 \\
\rowcolor{blue!5} &  {\bf \texttt{Sink-Aware}}  & \textbf{57.68} & \textbf{65.33} & \textbf{43.09} & 74.37 & \textbf{69.53} & \textbf{68.58} & 71.98 & 30.12 & \textbf{38.40} \\
\midrule
\multirow{4}{*}[-1.1ex]{0.50} & Wanda & 52.70 & 61.43 & 39.08 & 72.63 & 64.56 & \textbf{57.01} & \textbf{67.52} & 27.15 & 32.20 \\
\rowcolor{blue!5} & {\bf \texttt{Sink-Aware}}  & \cellcolor{gray!20}\textbf{53.18} & \textbf{62.16} & \textbf{41.38} & \textbf{73.18} & \textbf{65.27} & 55.88 & 67.18 & \textbf{27.95} & \textbf{32.45} \\
\cmidrule(lr){2-11}
& SparseGPT & 52.34 & \textbf{60.97} & \textbf{39.68} & 72.20 & 64.64 & \textbf{53.53} & 66.90 & \textbf{27.70} & \textbf{33.10} \\
\rowcolor{blue!5} &  {\bf \texttt{Sink-Aware}}  & \cellcolor{gray!20}\textbf{52.36} & 60.79 & 39.59 & \textbf{72.95} & \textbf{65.82} & 52.11 & \textbf{67.35} & 27.48 & 32.82 \\
\midrule
\multirow{4}{*}[-1.1ex]{0.75} & Wanda & 29.99 & \textbf{24.76} & 18.52 & 56.69 & 47.43 & 0.99 & 45.25 & 22.85 & 23.45 \\
\rowcolor{blue!5} &  {\bf \texttt{Sink-Aware}}  & \cellcolor{gray!20}\textbf{30.94} & 24.01 & \textbf{18.77} & \textbf{59.96} & \textbf{49.17} & \textbf{1.52} & \textbf{46.85} & \textbf{23.10} & \textbf{24.10} \\
\cmidrule(lr){2-11}
& SparseGPT & \textbf{32.57} & 28.60 & 20.99 & \textbf{61.75} & 50.04 & 1.52 & 48.20 & \textbf{23.90} & \textbf{25.55} \\
\rowcolor{blue!5} &  {\bf \texttt{Sink-Aware}}  & \cellcolor{gray!20}\textbf{32.57} & \textbf{28.93} & \textbf{21.08} & 60.12 & \textbf{51.07} & \textbf{1.90} & \textbf{48.70} & 23.55 & 25.20 \\
\bottomrule
\end{tabular}
}
\caption{Pruning results on {\bf LLaDA} across 8 benchmarks.} 
\label{tab:llada_results}
\end{table*}

\begin{table*}[t]
\centering
\small
\setlength{\tabcolsep}{3.5pt}
\renewcommand{\arraystretch}{1.08}
\resizebox{0.84\textwidth}{!}{
\begin{tabular}{c l >{\columncolor{gray!20}}c c c c c c c c c}
\toprule
\textbf{Sparsity} & \textbf{Method} & \cellcolor{gray!20}\textbf{Avg.} & \textbf{MMLU} & \textbf{ARC-C} & \textbf{PIQA} & \textbf{WinoG} & \textbf{GSM8K} & \textbf{HellaSwag} & \textbf{GPQA} & \textbf{RACE} \\
\midrule
\multirow{1}{*}{Dense} & \textbf{Base} & 60.94 & 68.25 & 55.20 & 74.27 & 67.72 & 67.50 & 73.30 & 36.60 & 44.70 \\
\midrule
\multirow{4}{*}[-1.1ex]{0.50} & Wanda & 51.58 & \textbf{60.90} & 41.64 & 68.12 & \textbf{60.38} & 49.13 & \textbf{65.40} & \textbf{30.85} & \textbf{36.20} \\
\rowcolor{blue!5} &  {\bf \texttt{Sink-Aware}}  & \cellcolor{gray!20}\textbf{51.68} & 60.84 & \textbf{42.83} & \textbf{68.82} & 59.75 & \textbf{49.66} & 65.15 & 30.52 & 35.85 \\
\cmidrule(lr){2-11}
& SparseGPT & 54.40 & 64.13 & 44.88 & \textbf{70.18} & 63.54 & 52.84 & \textbf{68.72} & 32.48 & 38.42 \\
\rowcolor{blue!5} &  {\bf \texttt{Sink-Aware}}  & \cellcolor{gray!20}\textbf{54.58} & \textbf{64.46} & \textbf{44.97} & \textbf{70.18} & \textbf{64.09} & \textbf{53.15} & 68.35 & \textbf{32.90} & \textbf{38.58} \\
\midrule
\multirow{4}{*}[-1.1ex]{0.75} & Wanda & 30.38 & 23.96 & 20.65 & \textbf{53.43} & 52.17 & 1.06 & 45.12 & 23.50 & \textbf{23.15} \\
\rowcolor{blue!5} &  {\bf \texttt{Sink-Aware}}  & \cellcolor{gray!20}\textbf{30.52} & \textbf{24.10} & \textbf{21.08} & 52.99 & \textbf{52.25} & \textbf{1.14} & \textbf{46.05} & \textbf{23.72} & 22.80 \\
\cmidrule(lr){2-11}
& SparseGPT & 31.72 & 26.92 & \textbf{22.18} & 54.52 & 51.30 & 1.44 & \textbf{48.20} & 24.30 & \textbf{24.90} \\
\rowcolor{blue!5} &  {\bf \texttt{Sink-Aware}}  & \cellcolor{gray!20}\textbf{31.83} & \textbf{28.02} & 21.25 & \textbf{54.73} & \textbf{51.93} & \textbf{1.59} & 47.90 & \textbf{24.65} & 24.55 \\
\bottomrule
\end{tabular}
}
\caption{Pruning results on {\bf Dream} across 8 benchmarks.} 
\label{tab:dream_results}
\end{table*}

\paragraph{Baselines and Implementation Details.}
We compare our method against several widely used pruning baselines, including Wanda~\citep{sun2023simple},
SparseGPT~\citep{frantar2023sparsegpt}, and magnitude-based pruning.
All baselines are implemented based on the open-source codebases of Wanda and SparseGPT for fair comparison.

For sink-aware pruning, we derive importance scores from attention maps aggregated across all layers and heads,
and apply the resulting pruning masks uniformly to all pruned layers, including both feed-forward and attention layers.
Importance scores are computed using a calibration set drawn from WikiText-2~\citep{merity2016pointer},
consisting of 128 randomly sampled sequences each truncated to a length of 2048 tokens.
The same calibration data is used for all baselines to ensure a controlled and fair comparison.

We evaluate both unstructured and structured pruning settings.
For unstructured pruning, we consider sparsity ratios of 25\%, 50\%, and 75\%.
For structured pruning, following prior work, we evaluate multiple pruning levels corresponding to
progressively removing structured components of the model.
All experiments are conducted under identical configurations across models and benchmarks,
and all results are reported using the same evaluation protocols.

\begin{table*}[t]
\centering
\small
\setlength{\tabcolsep}{3.5pt}
\renewcommand{\arraystretch}{1.08}
\resizebox{0.85\textwidth}{!}{
\begin{tabular}{c l >{\columncolor{gray!20}}c c c c c c c c c}
\toprule
\textbf{Sparsity} & \textbf{Method} & \cellcolor{gray!20}\textbf{Avg.} & \textbf{MMLU} & \textbf{ARC-C} & \textbf{PIQA} & \textbf{WinoG} & \textbf{GSM8K} & \textbf{HellaSwag} & \textbf{GPQA} & \textbf{RACE} \\
\midrule
\multirow{1}{*}{Dense} & \textbf{Base} & 58.59 & 64.07 & 50.43 & 73.61 & 68.90 & 65.73 & 74.43 & 31.75 & 39.87 \\
\midrule
\multirow{4}{*}[-1.1ex]{0.25} & Wanda & 58.09 & 63.30 & 51.37 & 74.81 & 68.19 & 62.47 & \textbf{73.68} & 31.20 & 39.72 \\
\rowcolor{blue!5} &  {\bf \texttt{Sink-Aware}}  & \cellcolor{gray!20}\textbf{58.22} & \textbf{63.41} & \textbf{51.54} & \textbf{74.97} & \textbf{68.27} & \textbf{62.70} & 73.55 & \textbf{31.30} & \textbf{40.05} \\
\cmidrule(lr){2-11}
& SparseGPT & 58.44 & \textbf{63.62} & \textbf{52.05} & \textbf{74.97} & 68.51 & 63.08 & \textbf{73.92} & 31.55 & \textbf{39.85} \\
\rowcolor{blue!5} &  {\bf \texttt{Sink-Aware}}  & \cellcolor{gray!20}\textbf{58.47} & 63.59 & \textbf{52.05} & \textbf{74.97} & \textbf{68.68} & \textbf{63.23} & 73.78 & \textbf{31.62} & 39.80 \\
\midrule
\multirow{4}{*}[-1.1ex]{0.50} & Wanda & 53.78 & \textbf{59.04} & 47.27 & 72.69 & 66.85 & 51.63 & \textbf{69.55} & 28.35 & \textbf{34.82} \\
\rowcolor{blue!5} &  {\bf \texttt{Sink-Aware}}  & \cellcolor{gray!20}\textbf{54.02} & 58.67 & \textbf{48.63} & \textbf{73.39} & \textbf{66.93} & \textbf{52.01} & 69.38 & \textbf{28.48} & 34.65 \\
\cmidrule(lr){2-11}
& SparseGPT & 53.91 & 59.20 & 48.46 & 72.63 & 65.19 & 52.24 & \textbf{69.20} & \textbf{28.92} & \textbf{35.45} \\
\rowcolor{blue!5} &  {\bf \texttt{Sink-Aware}}  & \cellcolor{gray!20}\textbf{54.10} & \textbf{59.33} & \textbf{49.15} & \textbf{73.12} & \textbf{65.43} & \textbf{52.69} & 69.15 & 28.70 & 35.20 \\
\midrule
\multirow{4}{*}[-1.1ex]{0.75} & Wanda & 31.03 & 25.87 & 19.62 & 59.09 & 49.09 & 1.14 & 46.80 & 23.15 & 23.50 \\
\rowcolor{blue!5} &  {\bf \texttt{Sink-Aware}}  & \cellcolor{gray!20}\textbf{32.89} & \textbf{27.20} & \textbf{24.91} & \textbf{61.37} & \textbf{51.93} & \textbf{1.29} & \textbf{48.30} & \textbf{23.80} & \textbf{24.35} \\
\cmidrule(lr){2-11}
& SparseGPT & \textbf{33.94} & \textbf{29.85} & \textbf{26.88} & 61.21 & \textbf{52.09} & \textbf{1.67} & \textbf{49.85} & 24.05 & \textbf{25.90} \\
\rowcolor{blue!5} &  {\bf \texttt{Sink-Aware}}  & \cellcolor{gray!20} 33.63 & 29.08 & 25.77 & \textbf{61.86} & 51.38 & 1.52 & 49.50 & \textbf{24.25} & 25.65 \\
\bottomrule
\end{tabular}
}
\caption{Pruning results on {\bf LLaDA1.5} across 8 benchmarks. }
\label{tab:llada15_results}
\vspace{-0.15in}
\end{table*}


\begin{table}[t]
\centering
\small
\setlength{\tabcolsep}{4pt}
\resizebox{0.45\textwidth}{!}{
\begin{tabularx}{\columnwidth}{lcccc}
\toprule
\textbf{Model} & \textbf{PIQA} & \textbf{WinoG} & \textbf{ARC-E} & \textbf{ARC-C} \\
\midrule
\textbf{LLaDA (Base)} & 0.7942 & 0.7388 & 0.7420 & 0.4437 \\
\midrule
\multicolumn{5}{l}{\textit{Structured Pruning 0.3}} \\
LLaDA-structure & 0.6834 & 0.6630 & 0.6907 & 0.3780 \\
\rowcolor{blue!5} {\bf \texttt{Sink-Aware}} (Ours)  & \textbf{0.6955} & \textbf{0.6740} & \textbf{0.7175} & \textbf{0.3820} \\
\midrule
\multicolumn{5}{l}{\textit{Structured Pruning 0.5}} \\
LLaDA-structure & 0.5898 & 0.5572 & 0.4853 & 0.2039 \\
\rowcolor{blue!5} {\bf \texttt{Sink-Aware}} (Ours)  & \textbf{0.6037} & \textbf{0.5724} & \textbf{0.5279} & \textbf{0.2362} \\
\bottomrule
\end{tabularx}
}
\vspace{-0.05in}
\caption{Structured pruning results on LLaDA.}
\label{tab:structured_pruning}
\vspace{-0.15in}
\end{table}

\begin{figure*}[t]
  \centering
  \includegraphics[width=\textwidth]{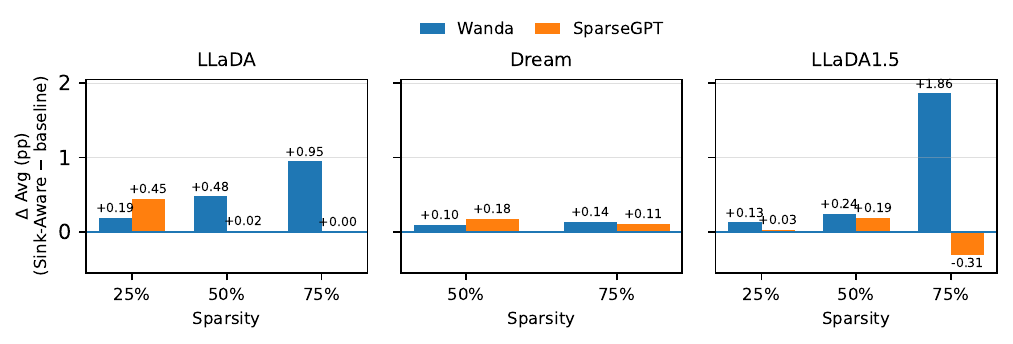}
  \vspace{-0.35in}
  \caption{\textbf{Delta vs. baseline on average accuracy.} Average accuracy change ($\Delta$Avg, in percentage points) from applying \texttt{Sink-Aware} pruning on top of each baseline (\textsc{Wanda} / \textsc{SparseGPT}) at different sparsity levels. Bars report $\Delta = \text{Avg}(\texttt{Sink-Aware}) - \text{Avg}(\text{baseline})$; positive values indicate improved performance retention after pruning. Results are shown for LLaDA, Dream, and LLaDA1.5 over 8 benchmarks.}
  \label{fig:delta_vs_baseline_avg}
  \vspace{-0.1in}
\end{figure*}

\subsection{Pruning Results}

\paragraph{Unstructured Pruning.}
Tables~\ref{tab:llada_results}--\ref{tab:mmada_results} report unstructured pruning results across all models and benchmarks. Sink-aware pruning consistently matches or surpasses the corresponding baselines at every sparsity level, demonstrating that accounting for the transient, denoising-step-dependent nature of attention sinks in DLMs yields more faithful compressed models than criteria designed for static AR-style sinks. The gains are most pronounced at moderate-to-high sparsity (50\%--75\%), where aggressive compression amplifies the cost of naively removing weights that serve unstable but functionally critical sink positions. At lower sparsity (25\%), the advantage narrows and is occasionally marginal, consistent with our observation that sink instability is not equally pronounced across all layers and model families, when the sparsity budget is generous, even standard importance criteria can avoid the most harmful removals. At 75\% sparsity, all methods degrade substantially, yet sink-aware pruning remains among the top-performing approaches, suggesting that discounting transient sinks provides a more principled lower bound on model utility under extreme compression. Importantly, the improvements are consistent across both Wanda and SparseGPT backbones, indicating that sink-aware masking acts as a general-purpose correction that complements existing importance scoring rather than being tied to a specific pruning criterion.

\paragraph{Structured Pruning.}
Table~\ref{tab:structured_pruning} presents structured pruning results on LLaDA. Sink-aware pruning consistently outperforms the structured baseline, with the margin growing at higher pruning ratios (0.3$\rightarrow$0.5). This trend is intuitive: structured pruning removes entire heads or layers at once, making each decision considerably more disruptive than removing individual weights, and therefore more sensitive to whether sink-critical components are retained. The growing improvement at higher ratios further reinforces our broader observation that sink-aware pruning is most valuable precisely when compression is aggressive, and the cost of misidentifying important structures is highest.  

\vspace{-0.05in}
\subsection{Visualization and Analysis}

Fig.~\ref{fig:delta_vs_baseline_avg} reveals a clear and consistent pattern: \texttt{Sink-Aware} pruning reliably improves over strong baselines across models and sparsity levels, with gains that grow as compression becomes more aggressive. The most notable improvements appear at 75\% sparsity, where preserving structurally critical parameters has the greatest impact, most evidently on LLaDA-1.5 under Wanda. At lower sparsity levels (25\% to 50\%), gains are smaller but remain broadly positive, suggesting that the sink-aware signal usefully complements existing importance criteria even in less demanding regimes.

\begin{figure*}[t]
  \centering
  \includegraphics[width=\textwidth]{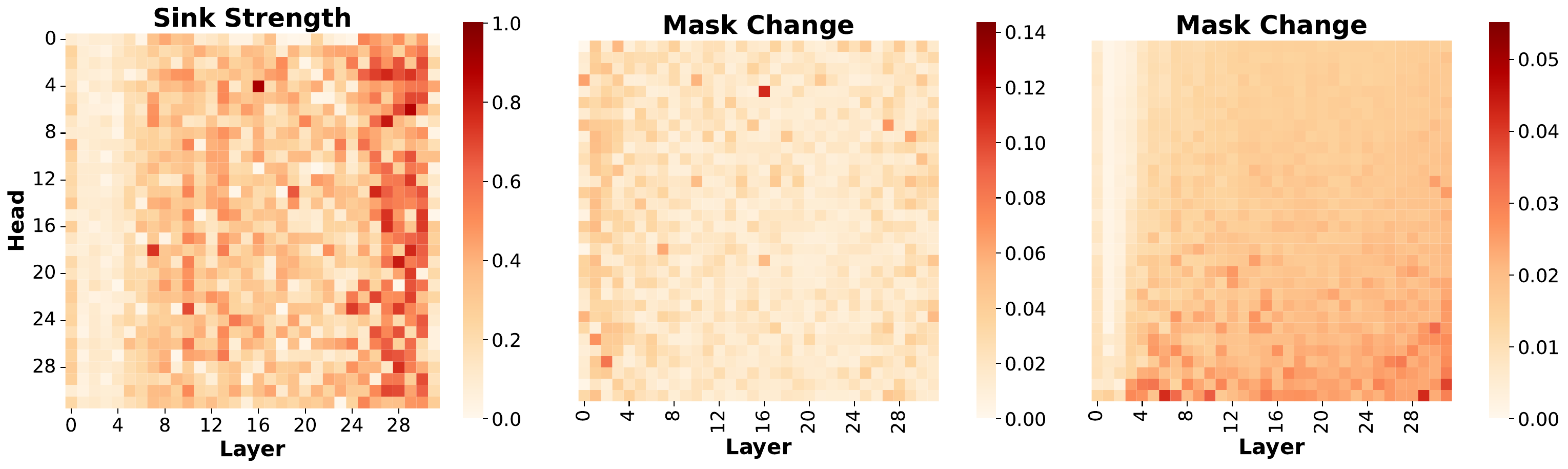}
  \vspace{-0.3in}
  \caption{\textbf{Sink strength and per-head mask disagreement (XOR) between baseline and sink-aware pruning.} 
The left subfigure shows the sink strength of the original LLaDA-8B model, measured as the average attention each head allocates to sink tokens. The center and right show the fraction of weights per head whose pruning decision differs (XOR) between the baseline and our method, using Wanda (center) and SparseGPT (right), respectively.}
\label{fig:mask_diff_xor}
\end{figure*}

\begin{figure*}[t]
  \centering
  \includegraphics[width=0.76\textwidth]{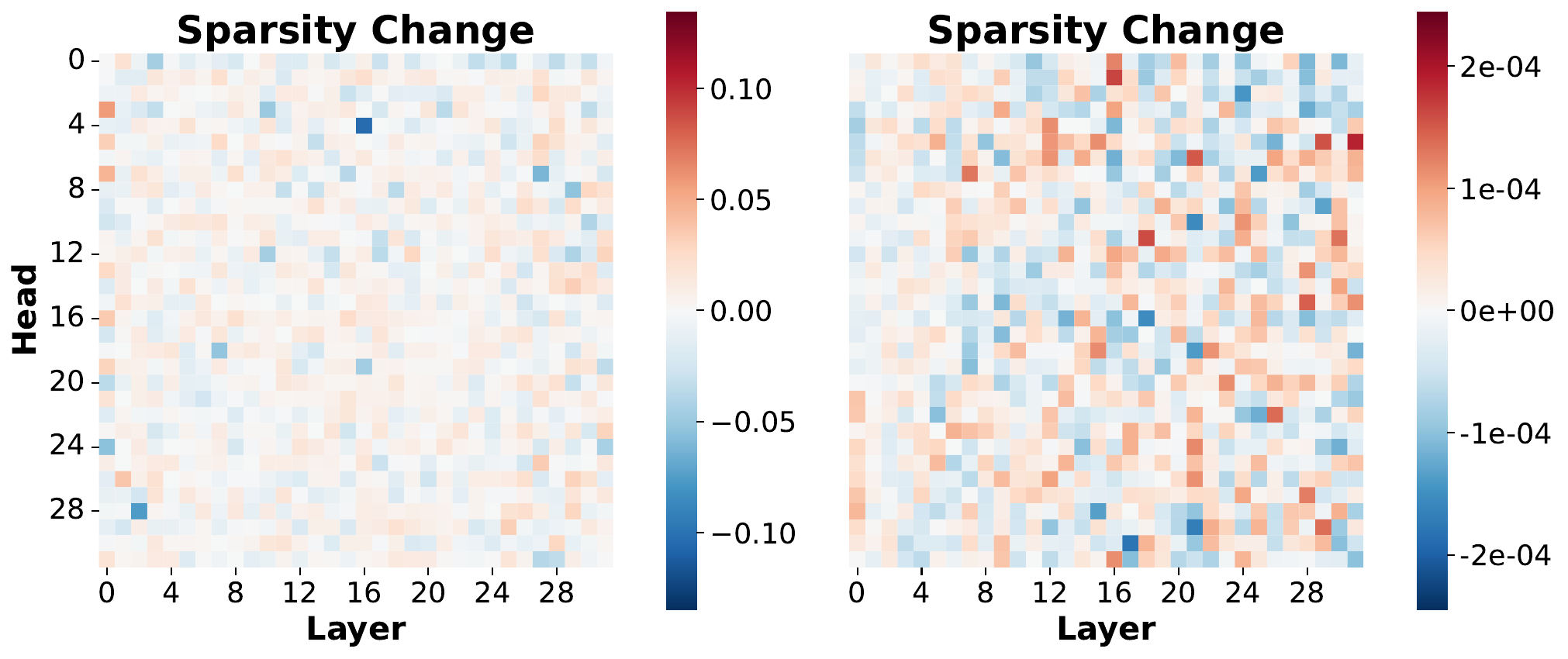}
  \vspace{-0.08in}
  \caption{\textbf{Per-head sparsity difference between baseline and sink-aware pruning.} 
The left and right subfigures show the signed per-head sparsity difference for Wanda and SparseGPT, respectively. Red indicates that the sink-aware variant prunes more aggressively in a given head, while blue indicates it preserves more weights. }
\label{fig:mask_diff_sparsity}
\end{figure*}

Fig.~\ref{fig:mask_diff_xor} and Fig.~\ref{fig:mask_diff_sparsity} show how \texttt{Sink-Aware} pruning alters head-level pruning decisions under a 50\% unstructured sparsity setting on LLaDA-8B. 
The result suggests a partial alignment between sink strength patterns and mask disagreement for Wanda, whereas the correspondence appears weaker for SparseGPT. 
The sparsity difference plot also indicates that Sink-Aware Pruning introduces smaller head-level pruning ratio changes for SparseGPT compared to Wanda.
\paragraph{Discussion.}
Our results suggest that sink behavior is generation-paradigm dependent.
In AR models, sinks are typically stable prefix anchors, in DLMs, sink positions shift across denoising steps and are often transient.
This explains why directly inheriting AR-style ``always keep sinks'' heuristics can be suboptimal for diffusion pruning.
By discounting unstable sinks, our {\texttt{Sink-Aware Pruning}} preserves a larger fraction of semantically useful parameters under the same sparsity budget, especially at moderate-to-high pruning levels.

We also observe that improvements are not uniform across all settings.
On some model-benchmark pairs (particularly in low-sparsity regimes), sink-aware variants are close to or slightly below strong baselines, indicating that sink instability is not equally pronounced in every layer, task, or model family.
This points to a practical trade-off: sink-aware pruning is most beneficial when compression is aggressive or when diffusion-time attention dynamics are highly non-stationary.

\vspace{-0.05in}
\section{Conclusion}

We revisited pruning for Diffusion Language Models from the perspective of attention-sink dynamics. Our analysis shows that sink positions in DLMs are substantially less stable than in AR models, which challenges the common sink-preservation heuristic inherited from AR pruning. Based on this observation, we proposed {\bf \texttt{Sink-Aware Pruning}}, a diffusion-oriented strategy that discounts unstable sinks when estimating pruning importance. Across multiple DLM families, benchmarks, and sparsity levels, the proposed method consistently improves or matches strong baselines while yielding a better quality-efficiency trade-off without retraining. These findings highlight that pruning rules should be aligned with generation dynamics rather than directly transferred across paradigms.

\section*{Limitations}
Our current study has several limitations.
First, sink statistics are estimated from a fixed calibration distribution, and distribution shift may reduce reliability.
Second, we mainly evaluate post-training pruning without recovery finetuning. Combining sink-aware scoring with lightweight post-pruning adaptation may further improve robustness.
Moreover, while we include one multimodal DLM, broader validation on larger multimodal and long-context settings is still needed.
Future work can explore layer-wise timestep-adaptive sink policies and joint optimization with quantization to further improve the quality-efficiency frontier.


\bibliography{custom}

\clearpage
\appendix
\section*{\Large{Appendix}}

\section{Results on Additional Models}
Table~\ref{tab:mmada_results} reports pruning results on MMaDA, a multimodal unified diffusion language model, complementing the main-paper results on LLaDA, Dream, and LLaDA-1.5.

\begin{table*}[b]
\centering
\small
\setlength{\tabcolsep}{3.5pt}
\renewcommand{\arraystretch}{1.08}
\resizebox{0.85\textwidth}{!}{
\begin{tabular}{c l >{\columncolor{gray!20}}c c c c c c c c c}
\toprule
\textbf{Sparsity} & \textbf{Method} & \textbf{Avg.} & \textbf{MMLU} & \textbf{ARC-C} & \textbf{PIQA} & \textbf{WinoG} & \textbf{GSM8K} & \textbf{HellaSwag} & \textbf{GPQA} & \textbf{RACE} \\
\midrule
\multirow{1}{*}{Dense} & \textbf{Base} & 35.63 & 33.69 & 24.32 & 60.50 & 50.83 & 9.48 & 52.50 & 25.20 & 28.50 \\
\midrule
\multirow{4}{*}[-1.1ex]{0.25}
& Wanda & \cellcolor{gray!20}\textbf{35.20} & 32.89 & 23.98 & \textbf{60.28} & \textbf{50.59} & 8.72 & \textbf{52.25} & \textbf{24.80} & \textbf{28.12} \\
\rowcolor{blue!5} & \bfseries\texttt{Sink-Aware} & \cellcolor{gray!20}{35.03} & \textbf{32.99} & \textbf{24.15} & 60.17 & 49.64 & \textbf{8.87} & 51.80 & 24.72 & 27.90 \\
\cmidrule(lr){2-11}
& SparseGPT & \textbf{35.48} & \textbf{33.30} & 24.57 & \textbf{60.39} & \textbf{50.91} & 9.10 & 52.05 & \textbf{25.15} & \textbf{28.35} \\
\rowcolor{blue!5} & \bfseries\texttt{Sink-Aware} & \cellcolor{gray!20}35.45 & \textbf{33.30} & \textbf{24.83} & 60.34 & 50.67 & \textbf{9.17} & \textbf{52.10} & 25.00 & 28.20 \\
\midrule
\multirow{4}{*}[-1.1ex]{0.50}
& Wanda & \textbf{33.91} & \textbf{31.14} & \textbf{23.12} & 59.79 & \textbf{51.14} & 5.38 & \textbf{49.72} & \textbf{24.18} & \textbf{26.80} \\
\rowcolor{blue!5} & \bfseries\texttt{Sink-Aware} & \cellcolor{gray!20}33.54 & 30.27 & 22.95 & \textbf{59.90} & 49.57 & \textbf{5.53} & 49.45 & 24.08 & 26.55 \\
\cmidrule(lr){2-11}
& SparseGPT & \textbf{34.05} & \textbf{31.33} & \textbf{23.81} & 59.36 & \textbf{50.04} & \textbf{5.84} & \textbf{50.25} & \textbf{24.55} & \textbf{27.20} \\
\rowcolor{blue!5} & \bfseries\texttt{Sink-Aware} & \cellcolor{gray!20}33.77 & 31.24 & 23.72 & \textbf{59.41} & 48.86 & 5.76 & 49.90 & 24.30 & 26.95 \\
\midrule
\multirow{4}{*}[-1.1ex]{0.75}
& Wanda & 29.85 & 23.93 & 19.37 & 54.79 & \textbf{50.12} & 0.76 & 42.55 & 23.10 & 24.15 \\
\rowcolor{blue!5} & \bfseries\texttt{Sink-Aware} & \cellcolor{gray!20}\textbf{30.28} & \textbf{24.82} & \textbf{20.65} & \textbf{55.33} & 49.64 & \textbf{0.91} & \textbf{42.90} & \textbf{23.55} & \textbf{24.40} \\
\cmidrule(lr){2-11}
& SparseGPT & \textbf{30.26} & 24.01 & \textbf{20.39} & 55.77 & 49.01 & 1.06 & \textbf{43.48} & \textbf{23.40} & \textbf{24.95} \\
\rowcolor{blue!5} & \bfseries\texttt{Sink-Aware} & \cellcolor{gray!20}30.19 & \textbf{24.51} & 18.86 & \textbf{55.82} & \textbf{50.04} & \textbf{1.14} & 43.15 & 23.28 & 24.70 \\
\bottomrule
\end{tabular}
}
\caption{Pruning results on {\bf MMaDA} across 8 benchmarks.}
\label{tab:mmada_results}
\end{table*}

\section{Evaluation Details}
For loglikelihood-based evaluations (MMLU, ARC-C, PIQA, WinoGrande, HellaSwag, RACE, and GPQA), we build on the official model implementations provided in the LLaDA\footnote{\url{https://github.com/ML-GSAI/LLaDA}} and Dream\footnote{\url{https://github.com/DreamLM/Dream}} GitHub repositories. For generation-based tasks (GSM8K), we leverage Fast-DLLM~\cite{wu2025fast} to enable fast inference. All evaluations are conducted using the \texttt{lm-evaluation-harness}~\cite{eval-harness} framework under identical configurations across all models and baselines.

For generation tasks, we use a generation length of 256 tokens, a block length of 256, and 256 denoising steps. For loglikelihood-based benchmarks, we follow the Monte Carlo estimation protocol with 128 samples across all tasks, except for MMLU where we use a single sample following the evaluation convention established in the official LLaDA codebase.

\section{Model Architecture Details}
All models use a decoder-only Transformer backbone as the denoising network (mask predictor) in discrete diffusion. For LLaDA, the architecture follows a LLaMA-style design with RMSNorm, SwiGLU feed-forward blocks, and RoPE positional encoding \cite{nie2025large}. Dream-7B adopts the same base configuration as Qwen2.5-7B \cite{ye2025dream,yang2024qwen2}. MMaDA-8B follows the LLaDA configuration for text generation and is initialized from an LLaDA-8B checkpoint \cite{yang2025mmada}. Table~\ref{tab:model-arch} shows the total number of trainable parameters, the number of Transformer layers $L$, and the model width $d_{\text{model}}$.

\begin{table}[!htbp]
  \centering
  \setlength{\tabcolsep}{7pt}
  \renewcommand{\arraystretch}{1.0}
  \resizebox{0.45\textwidth}{!}{
  \begin{tabular}{lccc}
    \toprule
    Model & Params & \#Layers & $d_{\text{model}}$ \\
    \midrule
    LLaDA-8B     & 8.02B & 32 & 4096 \\
    LLaDA-1.5-8B & 8.02B & 32 & 4096 \\
    Dream-7B     & 7B    & 28 & 3584 \\
    MMaDA-8B     & 8.02B    & 32 & 4096 \\
    \bottomrule
  \end{tabular}
  }
  \caption{Backbone configurations.}
  \label{tab:model-arch}
\end{table}

\section{Benchmark Descriptions}

We evaluate text generation and reasoning using a diverse suite of established benchmarks:

\begin{itemize}
  \item \textbf{MMLU}~\citep{hendrycks2020measuring} evaluates broad factual knowledge and reasoning across 57 academic subjects using multiple-choice questions, and is commonly reported in few-shot settings.
  
  \item \textbf{ARC-C}~\citep{clark2018think} (AI2 Reasoning Challenge, \emph{Challenge} split) consists of multiple-choice grade-school science questions designed to require non-trivial reasoning beyond surface pattern matching.
  
  \item \textbf{PIQA}~\citep{bisk2020piqa} tests physical commonsense reasoning via multiple-choice questions about everyday situations, selecting the more plausible solution among two candidate answers.
  
  \item \textbf{WinoGrande}~\citep{sakaguchi2021winogrande} measures commonsense coreference and pronoun resolution using adversarially filtered, multiple-choice Winograd-style sentence pairs.
  
  \item \textbf{HellaSwag}~\citep{zellers2019hellaswag} assesses commonsense inference by selecting the most plausible continuation of a given context from multiple candidate endings.
  
  \item \textbf{RACE}~\citep{lai-etal-2017-race} is a reading comprehension benchmark derived from English examinations, requiring multi-sentence reasoning over passages with multiple-choice questions.
  
  \item \textbf{GSM8K}~\citep{cobbe2021training} evaluates multi-step mathematical reasoning on grade-school word problems; performance is typically measured by exact match on the final numeric answer.
  
  \item \textbf{GPQA}~\citep{rein2024gpqa} is a challenging, domain-expert-written benchmark of graduate-level multiple-choice questions in biology, physics, and chemistry, designed to be difficult to solve via simple lookup.
\end{itemize}

\end{document}